\documentclass[letterpaper, 10 pt, conference]{ieeeconf}  

\IEEEoverridecommandlockouts                              

\usepackage{cite}
\usepackage{amsmath,amssymb,amsfonts}
\usepackage{algorithmic}
\usepackage{graphicx}
\usepackage{subcaption}
\usepackage{textcomp}
\usepackage{xcolor}
\usepackage{booktabs} 
\usepackage{multirow} 
\usepackage{amsmath}  
\usepackage{array}    
\usepackage{caption}  
\newcolumntype{C}{>{\centering\arraybackslash}p{2cm}} 
\usepackage{amssymb,amsmath,amsopn,amsfonts,colordvi}
\usepackage{bm}
\usepackage{booktabs}   
\usepackage{multirow}   
\usepackage{siunitx}    
\usepackage{dsfont}
\usepackage{graphicx}
\usepackage{color}
\usepackage{graphicx}
\usepackage{epstopdf}
\usepackage{float}
\usepackage{url}
\definecolor{orange}{rgb}{1,0.5,0}
\usepackage[colorlinks,linkcolor=orange,anchorcolor=red,citecolor=orange]{hyperref}
\newtheorem{remark}{Remark}[section]
\newtheorem{assumption}{Assumption}[section]

\newtheorem{theorem}{Theorem}[section]

\newcommand{\beq}{\begin{equation}}
\newcommand{\eeq}{\end{equation}}
\newcommand{\bea}{\begin{eqnarray}}
\newcommand{\eea}{\end{eqnarray}}
\newcommand{\bef}{\begin{flalign}}
\newcommand{\eef}{\end{flalign}}

\def\BibTeX{{\rm B\kern-.05em{\sc i\kern-.025em b}\kern-.08em
    T\kern-.1667em\lower.7ex\hbox{E}\kern-.125emX}}

\title{\LARGE \bf
Set-Valued Transformer Network for High-Emission Mobile Source Identification
}

\author{Yunning Cao, Lihong Pei, Jian Guo, Yang Cao$^{*}$, Yu Kang$^{*}$ and Yanlong Zhao
\thanks{*This work was supported by the National Natural Science Foundation of China (Grants 62033012 and 62103125).}
\thanks{*Corresponding authors}%
\thanks{Yunning Cao and Yang Cao are with the Department of Automation, University of Science and Technology of China, Hefei, 230026, China.
        {\tt\small Emails: cyn92@mail.ustc.edu.cn, forrest@ustc.edu.cn.}}%
\thanks{Lihong Pei and Yanlong Zhao are with the State Key Laboratory of Mathematical Sciences, Academy of Mathematics and Systems Science, Chinese Academy of Sciences, Beijing 100190, China, also with School of Mathematical Sciences, University of Chinese Academy of Sciences, Beijing 100049, China.
        {\tt\small Emails:  plh1225@amss.ac.cn, ylzhao@amss.ac.cn.}}%
\thanks{Jian Guo is with CAS AMSS-PolyU Joint Laboratory of Applied Mathematics, The Hong Kong Polytechnic University, Hong Kong, China.
        {\tt\small Email: j.guo@amss.ac.cn}}%
\thanks{Yu Kang is with the Department of Automation, University of Science and Technology of China, Hefei, 230026, China, with the Department of Automation, School of Electrical and Automation Engineering, Hefei University of Technology, and also with Anhui Province Key Laboratory of Intelligent Low-Carbon Information Technology and Equipment, University of Science and Technology of China, Hefei 230027, China.
        {\tt\small Email: kangduyu@ustc.edu.cn.}}%
}

\begin{document}

\maketitle

\begin{abstract}

Identifying high-emission vehicles is a crucial step in regulating urban pollution levels and formulating traffic emission reduction strategies. However, in practical monitoring data, the proportion of high-emission state data is significantly lower compared to normal emission states. This characteristic long-tailed distribution severely impedes the extraction of discriminative features for emission state identification during data mining. Furthermore, the highly nonlinear nature of vehicle emission states and the lack of relevant prior knowledge also pose significant challenges to the construction of identification models.To address the aforementioned issues, we propose a Set-Valued Transformer Network (SVTN) to achieve comprehensive learning of discriminative features from high-emission samples, thereby enhancing detection accuracy. Specifically, this model first employs the transformer to measure the temporal similarity of micro-trip condition variations, thus constructing a mapping rule that projects the original high-dimensional emission data into a low-dimensional feature space. Next, a set-valued identification algorithm is used to probabilistically model the relationship between the generated feature vectors and their labels, providing an accurate metric criterion for the classification algorithm. To validate the effectiveness of our proposed approach, we conducted extensive experiments on the diesel vehicle monitoring data of Hefei city in 2020. The results demonstrate that our method achieves a 9.5\% reduction in the missed detection rate for high-emission vehicles compared to the transformer-based baseline, highlighting its superior capability in accurately identifying high-emission mobile pollution sources.

\end{abstract}

\section{INTRODUCTION}

The rapid rise in vehicle numbers has significantly worsened air quality, making the accurate and efficient identification of high-emission vehicles a critical priority for environmental protection efforts.

\begin{figure}[tbp]
	\centering
	\begin{subfigure}[b]{0.5\linewidth} 
		\includegraphics[width=\linewidth,height=1.1in]{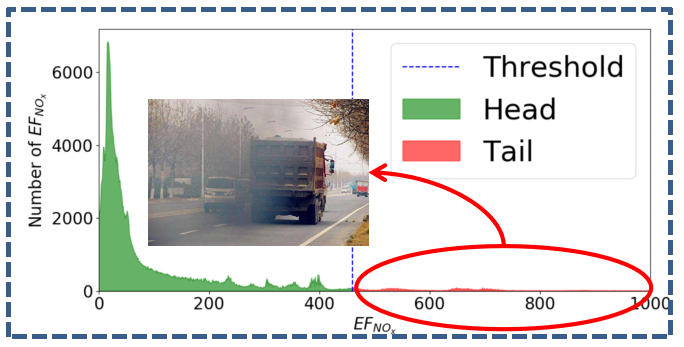}
		\caption{}
		\label{fig:1a}
	\end{subfigure}\hspace{-0.15cm}
	\begin{subfigure}[b]{0.5\linewidth}
		\includegraphics[width=\linewidth,height=1.1in]{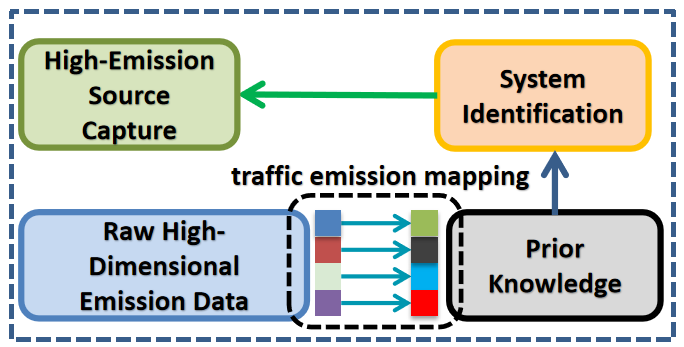}
		\caption{}
		\label{fig:1b}
	\end{subfigure}
	\caption{(a) Illustration of diesel vehicle emission distribution; (b) Illustration of the dual data-model driven approach for emission detection.}
	\label{fig:data-model}
\end{figure}

Rich monitoring data is crucial for intelligent excessive emission detection. Typically, emission data follows a long-tailed distribution, with normal states dominating while critical excessive emissions are severely underrepresented (Fig.~\ref{fig:1a}). This imbalance leads to significant pollution risks, making accurate detection vital to avoid both false positives and missed high-emission sources.

The vigorous promotion of digital intelligent supervision has effectively improved the efficiency of precise governance. Current methods for identifying excessive emission states can be mainly divided into two paradigms: data-driven and model-driven approaches. Data-driven methods capture underlying patterns from observed data with high representational accuracy, but they have inherent limitations when dealing with long-tailed distributed monitoring data—the extreme scarcity of excessive emission samples hinders sufficient learning of discriminative features, causing algorithms to be biased toward normal emission states and prone to misjudgment and missed detection of excessive emissions. The model-driven approach constructs models with clear physical significance by integrating domain knowledge and physical principles. It requires relatively low data inputs while maintaining strong interpretability. However, the highly nonlinear nature of complex driving conditions in mobile road sources and the lack of prior knowledge present significant challenges to system modeling.

Observations reveal that data-driven models excel at uncovering implicit knowledge rules from unknown data, thereby compensating for the limitations of model-driven algorithms in understanding complex systems. Meanwhile, model-driven approaches mitigate the over-reliance on data characteristics in emission identification by incorporating physical mechanisms and prior knowledge, thus enhancing the model's generalization capability and interpretability. As shown in Fig.~\ref{fig:1b}, the complementary strengths of these two paradigms provide a novel research direction for the accurate identification of high-emission mobile sources.

This paper constructs a transformer network based on the set-valued identification algorithm (Set-Valued Transformer Network, SVTN). This  framework maintains the powerful representational  capability of deep neural networks while incorporating identification algorithm constraints, significantly improving model robustness in small-sample  scenarios.  Specifically, this study reconstructs the classifier structure through decision boundary optimization theory, achieving systematic improvement  in binary classification performance. The key contributions include:

\begin{itemize}
    \item[1)] Innovatively combining the set-valued identification algorithm with the transformer network to fully learn discriminative features of high-emission samples, effectively reducing the misjudgment rate (false alarms) and missed detection rate (missed alarms) of excessive emission states.
    
    \item[2)] Employing the set-valued identification algorithm to probabilistically model the relationship between emission features represented by the transformer network and their labels, providing an accurate metric criterion for the classification algorithm.
    
    \item[3)] Experiments on diesel vehicle emission data in Hefei City demonstrate that SVTN exhibits significant advantages over conventional transformer algorithms in high-emission sample identification tasks, achieving a 9.5\% improvement in Recall and a 5.5\% increase in F1-score. These results validate the SVTN model's precise identification capability for high-emission states of mobile pollution sources.
\end{itemize}


\section{RELATED WORK}\label{Section 2}

The application of intelligent algorithms in engineering has grown rapidly in recent years. Methodologically, they fall into three categories: data-driven, model-driven, and data-model fusion approaches. Data-driven methods, empowered by deep learning's representational capacity, show significant advantages in complex scenarios like industrial process optimization and intelligent traffic management.
Kang \textit{et al.} \cite{KANG2021117877} use one-class classification and graph-based label propagation to address scarce positive labels. Pei \textit{et al.} \cite{10187171} propose a self-supervised GCN for vehicle emission clustering. Zhao \textit{et al.} \cite{Zhao2024ASL} develop a peak-sensitive microscopic emission estimation framework characterized by sequence-to-sequence learning for on-road vehicles. Xu \textit{et al.} \cite{xu2023high} propose a temporal optimization long short-term memory (LSTM) and adaptive dynamic threshold approach to identify heavy-duty high-emitters by using OBD data. The model-driven approach is based on physical mechanism models, and its excellent interpretability makes it indispensable in fields with high-reliability requirements such as aerospace and national defense security. Guo \textit{et al.} \cite{GUO2021479} design a distributed DRL with LSTM for UAV navigation in dynamic environments. Sun \textit{et al.} \cite{8946762} present resilient MPC against DoS attacks in CPS.
The data-model fusion method, by synergistically leveraging data-driven learning capabilities and prior knowledge of mechanistic models, has demonstrated unique advantages in complex engineering systems. THALER \textit{et al.} \cite{THALER202338125} develop a novel model predictive control strategy for a renewable microgrid with seasonal hydrogen storage. Tu \textit{et al.} \cite{TU2023120289} proposes two new frameworks to integrate physics-based models with machine learning to achieve high-precision modeling for LiBs. 

With the advancement of Industry~4.0 and the digital wave, set-valued systems have gradually matured as an innovative solution. In recent years, numerous innovative identification methods have been successfully applied to set-valued systems \cite{BI20143220, 10383964, GUO20133396}. With the deepening of theoretical research the applications of set-valued systems continue to expand in many fields, demonstrating unique advantages in solving complex system problems. Li \textit{et al.} \cite{Li2022} study the problem of lithology identification based on the set-valued method. Bi \textit{et al.} \cite{6640899} propose a set-valued analytical approach for genome-wide association studies (GWAS) of complex diseases. In parallel, adaptive estimation techniques for MIMO systems under unknown directions and parameter bounds have also made notable progress \cite{10783042}, providing valuable inspiration for learning-based identification under uncertainty.

\begin{figure*}[tbp]
    \centering 
    \includegraphics[width=0.7\linewidth]{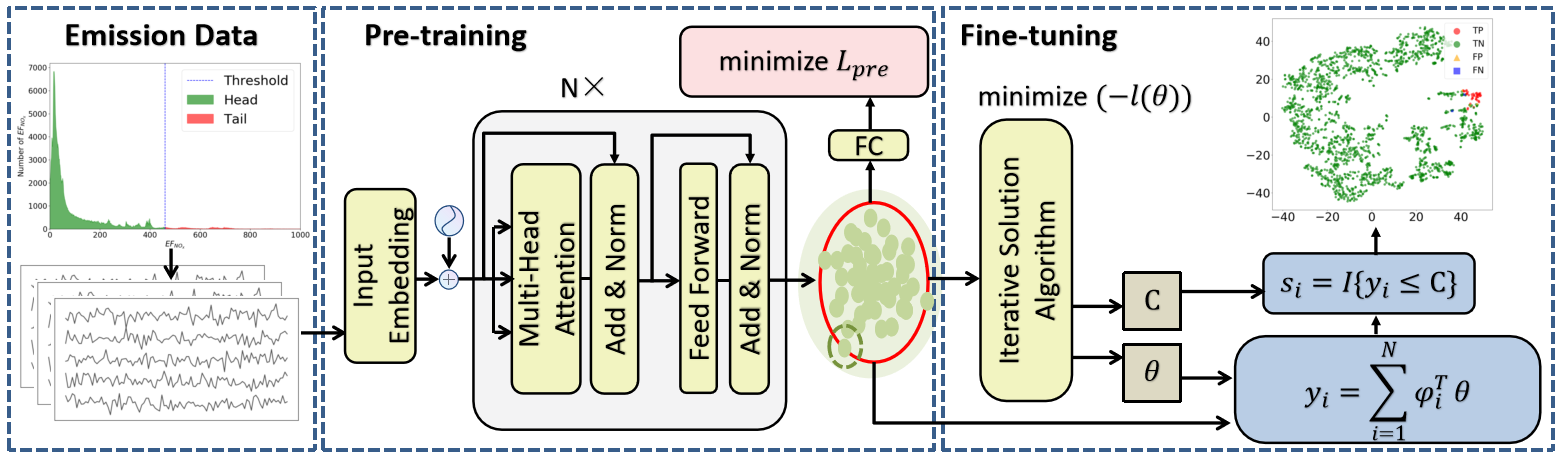} 
    \caption{Illustration of proposed SVTN. In the pre-training stage, the transformer network maps the original emission data to the feature space and constructs prior rules. In the fine-tuning stage, the set-valued algorithm explicitly models the binary classifier to obtain the final excessive emission label.} 
    \label{fig:10} 
\end{figure*}

\section{BINARY CLASSIFICATION USING SET-VALUED IDENTIFICATION}\label{Section 3}

To address the long-tailed distribution phenomenon in actual emission data, this paper proposes a SVTN model to achieve accurate identification of high-emission mobile sources. As shown in Fig.~\ref{fig:10}, our model consists of two stages: pre-training using the transformer encoder and fine-tuning with the set-valued identification algorithm.

To tackle the highly nonlinear characteristics exhibited by road mobile sources under complex driving conditions, this study employs a transformer architecture during the pre-training phase to quantify temporal similarities in micro-trip operational patterns. Specifically, we develop an emission pattern probability assessment model based on similarity metrics between vehicle operating data and characteristic emission profiles. The model dynamically evaluates the probability of a driving segment belonging to the normal emission regime through spatial distance computation, where closer proximity to reference features indicates higher probability of normal operation. The pre-training objective $L_{pre}$ employs the binary cross-entropy with logits loss (BCEWithLogits), defined as:
\begin{equation}
\small
L_{\text{pre}}\! \!=\!\! -\frac{1}{N}\sum_{i=1}^{N} \Bigl[z_i\log(\sigma(\hat{z}_i)) + (1-z_i)\log(1-\sigma(\hat{z}_i))\Bigr].
\end{equation}
where $\hat{z}_i \in \mathbb{R}$ denotes the model's logit output for the $i$-th sample, $z_i \in \{0,1\}$ represents the ground truth label, $\sigma(x) = 1/(1+e^{-x})$ is the sigmoid activation function, and $N$ indicates the batch size. Through backpropagation with this loss formulation, our experimental results demonstrate that the trained transformer model effectively captures emission behavior characteristics across diverse operating conditions, providing physically meaningful representations for downstream emission state classification tasks.

In binary classification tasks, transformer models can generate feature vectors \({\varphi}_i = \mathcal{T}(X_i)\) from the input data \(X_i \in \mathbb{R}^p\). These feature vectors can be directly used in binary classification tasks.

However, treating \({\varphi}_i\) as a deterministic feature vector fails to account for the noise and uncertainty introduced by the transformer model and sample observations. The feature vector \({\varphi}_i\) is not a fixed point but rather a sample from a distribution that reflects the inherent uncertainty in the learning process. To address this, we propose modeling the relationship between the generated feature vectors and the labels probabilistically, rather than deterministically. We introduce the distributional model by assuming that the feature vector \({\varphi}_i\) follows a probability distribution, and the label \(s_i\) is dependent on \({\varphi}_i\). Specifically, we model the probability distribution of \(s_i\) given \(X_i\) and \({\varphi}_i\) as:
\begin{equation}
	s_i \sim p(s_i | X_i, {\varphi}_i), \quad X_i \sim P_X,
\end{equation}
where \({\varphi}_i = \mathcal{T}(X_i)\) and \(p(s_i | X_i, {\varphi}_i)\) is the conditional probability distribution of the label \(s_i\).
\begin{remark}
	This approach considers the uncertainty in feature generation, where \( {\varphi}_i \) is a random variable influenced by \( X_i \) and transformer-induced noise. Thus, instead of a deterministic \( X_i \)-\( s_i \) relationship, \( s_i \) is treated as a conditional random variable.
\end{remark}

To simplify the modeling, we first parameterize the conditional distribution \(p(y_i | X_i, {\varphi}_i)\) with a parameter \(\theta\), and then use the high-dimensional generalized linear models (GLMs) with binary outcomes \cite{cai2023statistical}. Specifically, we model \(s_i\) as a Bernoulli random variable with a probability governed by a linear transformation of the input data:
\begin{equation}\label{GLM}
	s_i | X_i \sim \text{Bernoulli}(F(\varphi_i^T \theta)),
\end{equation}
where \(F(\cdot)\) is the link function, \(\theta \in \mathbb{R}^n\) is a regression vector, and \({\varphi}_i = \mathcal{T}(X_i)\). This parameterization allows us to estimate the distribution and  further extends the general framework to handle noisy data and uncertain feature vectors effectively.

To solve the problem (\ref{GLM}), we first transform the original GLM problem into a set-valued identification problem. Specifically, we generalize the link functions by considering a class induced by a latent variable model, where the auxiliary random variable is defined as:
\begin{equation}
	\begin{split}
		y_i &= \varphi_i^T \theta - C + d_i, \\
		\text{where } d_i &\sim P_{d_i}, \quad 1 \leq i \leq N.
	\end{split}
	\label{set_valued1}
\end{equation}
Then, the observed binary outcome variable can be expressed as  
\begin{equation}
	s_i = \mathbb{I}(y_i \leq 0).
	\label{set_valued2}
\end{equation}
\begin{assumption}\label{ass:A_positive_definite}
	The matrix \( A = \sum_{i=1}^{N} \varphi_i \varphi_i^T \) is positive definite.
\end{assumption}
\begin{remark}
    The positive definiteness condition specified in Assumption \ref{ass:A_positive_definite} characterizes the persistent excitation requirements for parameter estimation when using batch-collected data, a standard prerequisite in system identification theory \cite{Ljung1998}.
\end{remark}

The problem of interest is to estimate the parameter \(\theta\) using the binary-valued observations \( \mathcal{O}_N = \{s_1, s_2, \ldots, s_N\} \) and input data \( \mathcal{I}_N = \{\varphi_1, \varphi_2, \ldots, \varphi_N\} \).
Which is equivalent to the binary GLM (\ref{GLM}) with \( F(\cdot) \) being the cumulative distribution function (CDF) of \( d_i \).  The problem (\ref{set_valued1})-(\ref{set_valued2}) is indeed the set-valued identification problem. 

In the fine-tuning stage, this study fully utilizes the prior knowledge of emission states obtained by pre-training and employs a set-valued estimation-based system identification algorithm to iteratively solve the model (\ref{set_valued1})-(\ref{set_valued2}). By adopting the minimization of the log-likelihood function $-l(\theta)$ as the optimization objective, the joint estimation of the system's unknown parameter $\theta$ and the decision threshold $C$ is achieved. This method not only ensures the global convergence of parameter identification but also effectively enhances the model's generalization performance through the incorporation of prior knowledge, ultimately realizing accurate identification of emission states.

\subsection{Maximum Likelihood Estimation and Its Convergence}
Given the input dataset 
$
\mathcal{I}_N = \{ \varphi_1, \varphi_2, \dots, \varphi_N \}
$
and the binary label dataset 
$
\mathcal{O}_N = \{ s_1, s_2, \dots, s_N \},
$
this section presents parameter estimation based on the Maximum Likelihood Estimation (MLE) criterion, along with properties of the strong consistency and asymptotic normality of the MLE. We consider the system (\ref{set_valued1})-(\ref{set_valued2}), for any \(i \leqslant N\). Under the condition of the input data \(\varphi_{i}\) and the parameter \(\theta\), the corresponding probabilities of observation \(s_{i} = 1\) and \(s_{i} = 0\) are as follows:
\begin{equation}
	\begin{aligned}
		P\{s_i = 1 | \varphi_i, \theta\}
		&= P\{d_i \leq C - \varphi_i^T\theta | \varphi_i, \theta\} \\
		&= F(C - \varphi_i^T\theta),
	\end{aligned}
	\label{eq:s1}
\end{equation}

\begin{equation}
	\begin{aligned}
		P\{s_i = 0 | \varphi_i, \theta\} &= 1 - P\{s_i = 1 | \varphi_i, \theta\} \\
		&= 1 - F(C - \varphi_i^T\theta).
	\end{aligned}
	\label{eq:s0}
\end{equation}
Using the preceding conditional probability framework, we formulate the likelihood function:
\begin{equation}
	\small
	\begin{aligned}
		&L(\theta) = P\{\mathcal{O}_N | \mathcal{I}_N, \theta\} = \prod_{i=1}^N P\{s_i | \varphi_i, \theta\} \\
		&= \prod_{\{i: s_i = 1\}} F(C - \varphi_i^T\theta) \cdot \prod_{\{i: s_i = 0\}} \big{(}1 - F(C - \varphi_i^T\theta)\big{)}.
	\end{aligned}
	\label{eq:likelihood}
\end{equation}
which captures the complete observation probability of $\mathcal{O}_N$ conditioned on the input dataset $\mathcal{I}_N$.


The corresponding ML estimate is the parameter that maximizes the log-likelihood function:
\begin{equation}
	\hat{\theta} = \arg \max_{\theta} l(\theta) = \arg \max_{\theta}\log(L(\theta)).
	\label{eq:ml_estimate}
\end{equation}
\begin{theorem}
There exists a sequence of random variables $\{\hat{\theta}_N\}$ and a random integer $n_2$ such that
	\begin{equation}
		P \big( s_N(\hat{\theta}_N) = 0, \ \forall N \geq n_2 \big) = 1,
	\end{equation}
	and $\hat{\theta}_N$ converges almost surely to $\theta$, i.e.,
	\begin{equation}\label{ascon}
		\hat{\theta}_N \to \theta, \quad a.s.
	\end{equation}
	Furthermore, with $I(\cdot)$ being the Fisher information matrix, the MLE is asymptotically normal, that is,
	\begin{equation}\label{asymp}
		\sqrt{N} (\hat{\theta}_N - \theta) \xrightarrow{d} \mathcal{N}(0, I^{-1}(\theta)).
	\end{equation}
\end{theorem}
\noindent\textbf{Proof.} 
Since the noise follows a normal distribution, it is easy to verify that the conditions of Theorem 4.17 in the textbook \cite{shao2008mathematical} are satisfied. Therefore, based on this theorem, the MLE (\ref{eq:ml_estimate})  in our setting exhibits the properties (\ref{ascon}) and (\ref{asymp}).
\begin{remark}
    While the log-likelihood $l_N(\theta)$ and maximum likelihood estimate $\hat{\theta}_N$ explicitly indicate their dependence on $N$ observations, we adopt the simplified notations $l(\theta)$ and $\hat{\theta}$ throughout this paper when the context permits unambiguous interpretation.
\end{remark}
\subsection{Iterative Solution Algorithm}
This section derives an iterative algorithm for the maximum likelihood estimate (MLE) from \( N \) observations and provides its exponential convergence rate. The algorithm is based on the EM method. At the \(t\)-th iteration, with estimate \(\hat{\theta}_{t}\), the EM algorithm constructs the function \(l(\theta | \hat{\theta}_{t})\), which satisfies the following two properties:

(i) \(l(\theta | \hat{\theta}_{t}) \leqslant l(\theta)\) holds for all \(\theta\),

(ii) \(l(\hat{\theta}_{t} | \hat{\theta}_{t}) = l(\hat{\theta}_{t})\).

Then, calculate \(\hat{\theta}_{t+1}\) that maximizes the function \(l(\theta | \hat{\theta}_{t})\) as the estimate of the \((t+1)\)-th iteration. We can see that
\begin{align}
    l(\hat{\theta}_{t+1}) &\geqslant l(\hat{\theta}_{t+1} | \hat{\theta}_{t}) = \max l(\theta | \hat{\theta}_{t}) \nonumber \\
    &\geqslant l(\hat{\theta}_{t} | \hat{\theta}_{t}) = l(\hat{\theta}_{t}).
\end{align}
The algorithm ensures monotonic log-likelihood growth, providing strong robustness. The EM procedure alternates between: constructing \(l(\theta | \hat{\theta}_{t})\) (E-step) and maximizing it (M-step).

In the binary-valued model framework, the E-step yields a quadratic representation of \(l(\theta | \hat{\theta}_{t})\):
\begin{small}
\begin{equation}
	\begin{aligned}
	&	l(\theta|\hat{\theta}_t) = -\frac{1}{2}\theta^T\left(\sum_{i=1}^N \varphi_i \varphi_i^T\right)\theta \\
		&\quad + \left[\left(\sum_{i=1}^N \varphi_i \varphi_i^T\right)\hat{\theta}_t - \left(\sum_{i=1}^N \varphi_i \cdot f(C-\varphi_i^T\hat{\theta}_t) \right. \right. \\
		&\quad \left. \left. \times \left[ \frac{I_{[s_i=1]}}{F(C-\varphi_i^T\hat{\theta}_t)} - \frac{I_{[s_i=0]}}{1-F(C-\varphi_i^T\hat{\theta}_t)} \right] \right) \right]^T \theta \\
		&\quad + l_1(\hat{\theta}_t).
	\end{aligned}
\end{equation}
\end{small}

Under Assumption \ref{ass:A_positive_definite}, the iterative algorithm is as follows:
\beq
\begin{aligned}
		&\hat{\theta}_{t+1} = \arg\max_{\theta} l(\theta|\hat{\theta}_t) \\
		&= \hat{\theta}_t - \left( \sum_{i=1}^{N} \varphi_i \varphi_i^T \right)^{-1}\times \left( \sum_{i=1}^{N} \varphi_i f(C-\varphi_i^T \hat{\theta}_t)\right.\\
		 &\left.\left[ \frac{I_{[s_i=1]}}{F(C-\varphi_i^T \hat{\theta}_t)} - \frac{I_{[s_i=0]}}{1-F(C-\varphi_i^T \hat{\theta}_t)} \right] \right).\label{algo}
\end{aligned}
\eeq

The above algorithm exhibits the following exponential convergence.
\begin{theorem}[Exponential convergence] Under assumption \ref{ass:A_positive_definite}, the iterative algorithm (\ref{algo}) satisfies
\begin{equation}\label{eq:convergence-rate}
	\|\hat{\theta}_t - \hat{\theta}\| \leq \sqrt{\frac{Q_1}{\lambda_{\min}(A)}} \cdot \frac{\sqrt{(1 - \epsilon)^t}}{1 - \sqrt{(1 - \epsilon)}}
\end{equation}
where $\hat{\theta}$ represents the Maximum Likelihood Estimator (MLE) as given by Eq. (\ref{eq:ml_estimate}), $\|\cdot\|$ denotes the Euclidean norm, $A = \sum_{k=1}^N \phi_k \phi_k^T$ is the sum of outer products of the feature vectors, $\lambda_{\min}(A)$ is the smallest eigenvalue of matrix $A$, and $Q_1 = (1 - \epsilon)^{-1} (\hat{\theta}_2 - \hat{\theta}_1)^T A (\hat{\theta}_2 - \hat{\theta}_1)$ represents a quantity derived from the difference between successive iterates of the algorithm. 
\end{theorem}

\subsection{Variations of the Set-Valued Transformer Network}

In addition to simple linear relationships, the system may also exhibit nonlinear relationships. Inspired by the set-valued model, we define \( k(\varphi_{ij}) = [(\varphi_{ij})^1, (\varphi_{ij})^2, \cdots, (\varphi_{ij})^{k}] \) as the power sequence of \(\varphi_{ij}\) from the 1st to the \( k \)-th power.Let $\varphi_i^k = [k(\varphi_{i1}), k(\varphi_{i2}), \cdots, k(\varphi_{iN})] \in \mathbb{R}^{Nk}$, where the model using $\varphi_i^k$ as the new feature vector is referred to as SVTN($k$), a variation of the set-valued  transformer method.

For example, SVTN(3) has the feature vector \([\varphi_{i1}, (\varphi_{i1})^2, (\varphi_{i1})^3, \cdots, \varphi_{iN}, (\varphi_{iN})^2, (\varphi_{iN})^3]^T\). In this study, we select SVTN(1), SVTN(3), SVTN(5), and SVTN(7) for experiments.

\section{EXPERIMENTS}\label{Section 4}

This section presents a comprehensive evaluation of SVTN classification approach through comparative experiments with three benchmark methods: Transformer architecture, Set-Valued (SV) classification, and Random Forest (RF). The performance assessment employs critical classification metrics including recall rate and f1-score to systematically validate SVTN's effectiveness in emission sample identification.


\subsection{Data Preprocessing}

The On-Board Diagnostics(OBD) system provides real-time monitoring of NO\textsubscript{x} ($c_{NO_x}$) in parts per million (ppm), exhaust mass flow rate ($Q_{exh}$) in kg/h, $EnT$ and $EnS$ are the instantaneous net output torque (Nm) and engine speed (rpm) of the engine respectively. Based on the NOx concentration-to-emission ratio calculation formula, these parameters are integrated into specific emission calculations:
\begin{equation}
	EF_{NO_x} = \frac{0.001587 \times c_{NO_x} \times Q_{exh}}{\pi \times EnT \times EnS /1.08 \times 10^6},
\end{equation}
where $EF_{NO_x}$ represents the specific NO\textsubscript{x} emission in g/kWh. According to China VI emission standards for heavy-duty diesel vehicles (GB17691-2018), the transient NO\textsubscript{x} limit is defined as 460 mg/kWh, which serves as the classification threshold in this study. The dataset comprises 10,000 samples with severe class imbalance (Class~1: 2.28\%; Class~0: 97.72\%). We applied stratified 8:2 train--test splitting to maintain distribution parity, ensuring experimental reliability.
\subsection{Evaluation Metrics}
Two metrics are employed to assess model performance:
\begin{equation}
	\begin{aligned}
		&\text{Recall} (R) = \tfrac{N_{\mathrm{tp}}}{N_{\mathrm{tp}} + N_{\mathrm{fn}}}, \\
		&\text{F1} = 2\tfrac{P \cdot R}{P + R}, \\
	\end{aligned}
\end{equation}
where $N_{\text{tp}}$ denotes correctly identified high-emission samples, 
$N_{\text{fn}}$ denotes high-emission samples misclassified as normal emissions and $P$ is precision. To achieve a perfect model, both recall and f1-score need to be maximized, ensuring accurate identification of high-emission samples while avoiding misclassification of normal samples as high-emission. In actual monitoring, since high-emission samples are of primary concern, the recall rate and f1 score mentioned refer specifically to the metrics for high-emission samples.
\subsection{Experimental results} 

\begin{figure}[tbp]
	\centering
	\begin{subfigure}[b]{0.49\linewidth} 
		\includegraphics[width=\linewidth,height=1.2in]{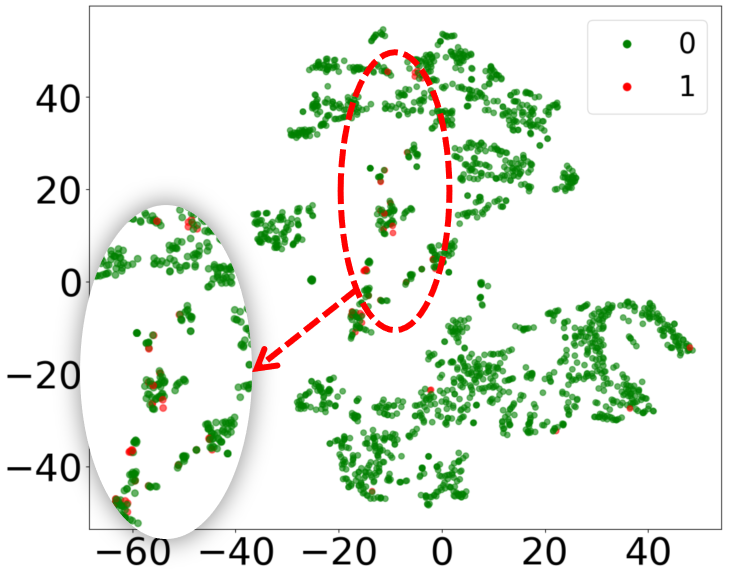}
		\caption{}
	\end{subfigure}
	\hfill 
	\begin{subfigure}[b]{0.49\linewidth}
		\includegraphics[width=\linewidth,height=1.2in]{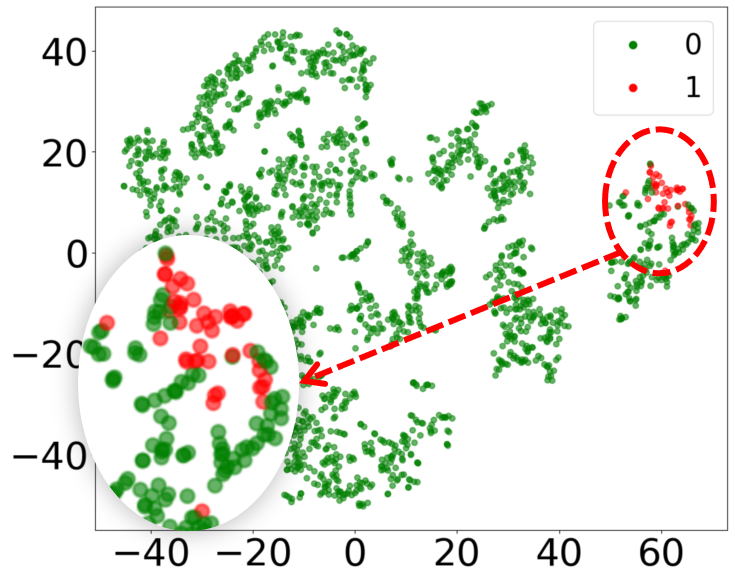}
		\caption{}
	\end{subfigure}
	\caption{Visualization of original data and features extracted by transformer}
	\label{fig:tsne}
\end{figure}

The original data distribution and the data distribution after feature extraction by the transformer are shown in Fig.~\ref{fig:tsne}. The high-dimensional 
data are reduced to two-dimensional data through t-SNE (t-distributed Stochastic Neighbor Embedding) \cite{JMLR:v9:vandermaaten08a}. It can be observed that the original data exhibits severe class imbalance, with high-emission samples (red) being significantly outnumbered by normal samples (green). The feature space learned by the transformer network demonstrates improved cluster cohesion and inter-class separation compared to the original distribution.

To reduce the randomness associated with single experiments, we conducted 20 repeated experiments to compare the SVTN(1) method with the traditional Transformer, SV(5), and RF methods. Tab.~\ref{tab:baselines} provides a detailed summary of the means and variances of the performance metrics for these four methods over the 20 experiments. The results show that the SVTN(1) method achieves a recall rate of 86.47\% and an f1 score of 89.16\%, both of which are significantly higher than those of the other methods. This indicates its superior balance between precision and recall, and highlights its accuracy and reliability in identifying high-emission samples.
\begin{table}[tbp]
    \centering
    \caption{Comparison of performance metrics for four methods} 
    \scriptsize
    \setlength{\tabcolsep}{4pt}
    \begin{tabular}{@{} l *{4}{S[table-format=2.2(3)]} @{}}
        \toprule
        \multicolumn{1}{c}{\textbf{Metrics}} & \multicolumn{4}{c}{\textbf{Methods}} \\
        \cmidrule(lr){1-1} \cmidrule(lr){2-5}
        & {SVTN(1)} & {Transformer} & {SV(5)} & {RF} \\
        \midrule
        Recall$ (\uparrow)$ & $\bm{86.47 \pm 4.56}$ & 76.96 \pm 6.87 & 81.52(692) & 80.98(387) \\
        F1-score$ (\uparrow)$ & $ \bm{89.16 \pm 3.42}$ & 83.71(436) & 87.00(461) & 88.40(260) \\
        \bottomrule
    \end{tabular}
    \label{tab:baselines}
\end{table}
\begin{figure}[!tbp]
    \centering
    \begin{subfigure}[b]{0.49\linewidth}
        \includegraphics[width=\linewidth,height=1.1in]{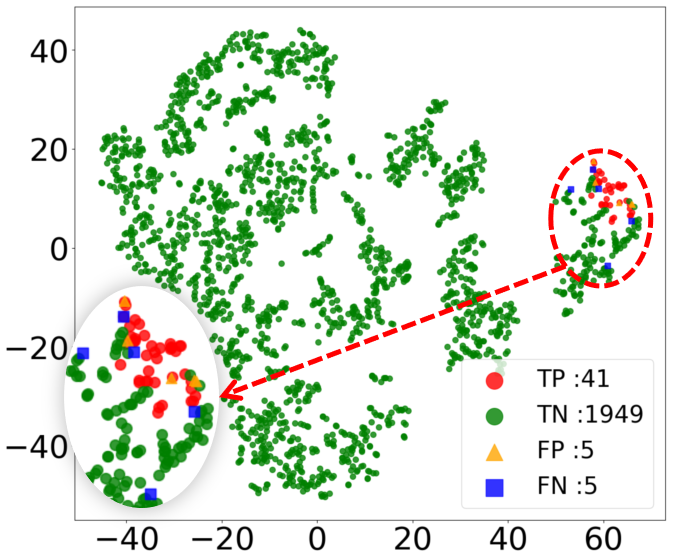}
        \caption{SVTN(1)}
    \end{subfigure}
    \hfill 
    \begin{subfigure}[b]{0.49\linewidth}
        \includegraphics[width=\linewidth,height=1.1in]{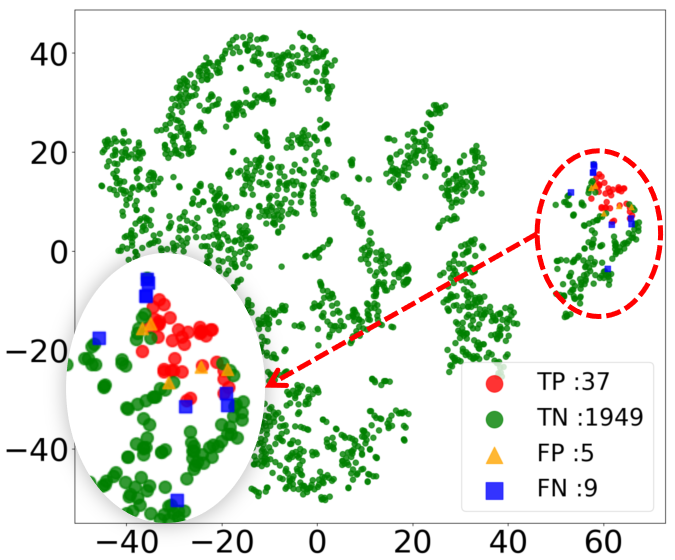}
        \caption{Transformer}
    \end{subfigure}
    
    \bigskip 
    \begin{subfigure}[b]{0.49\linewidth}
        \includegraphics[width=\linewidth,height=1.1in]{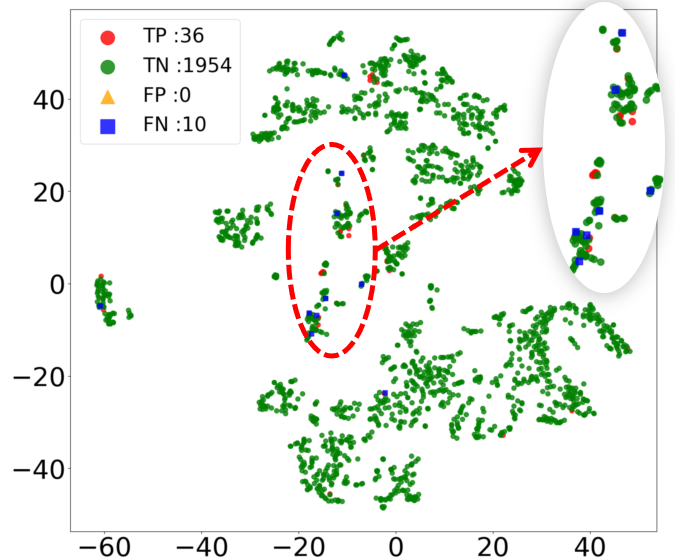}
        \caption{SV(5)}
    \end{subfigure}
    \hfill 
    \begin{subfigure}[b]{0.49\linewidth}
        \includegraphics[width=\linewidth,height=1.1in]{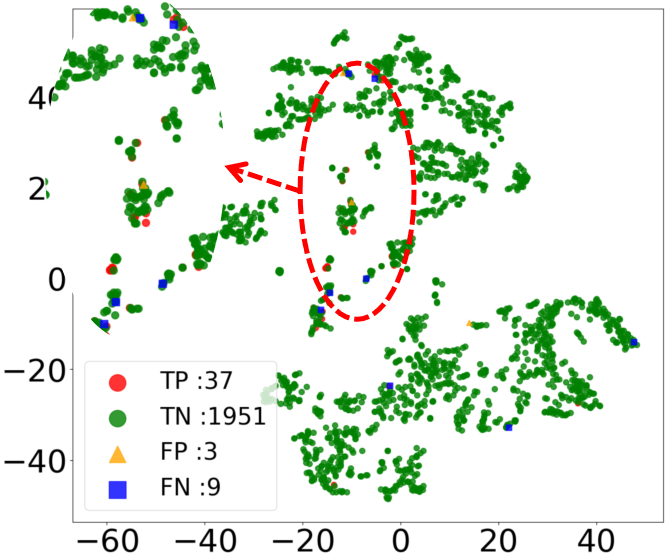}
        \caption{RF}
    \end{subfigure}
    \caption{Visualization of classification results}
    \label{fig:vcr}
\end{figure}

\begin{figure}[tbp]
    \centering
    \newcommand{\imageheight}{0.49\textheight}
    \begin{subfigure}[b]{0.49\linewidth}
        \includegraphics[width=\linewidth,height=1.1in]{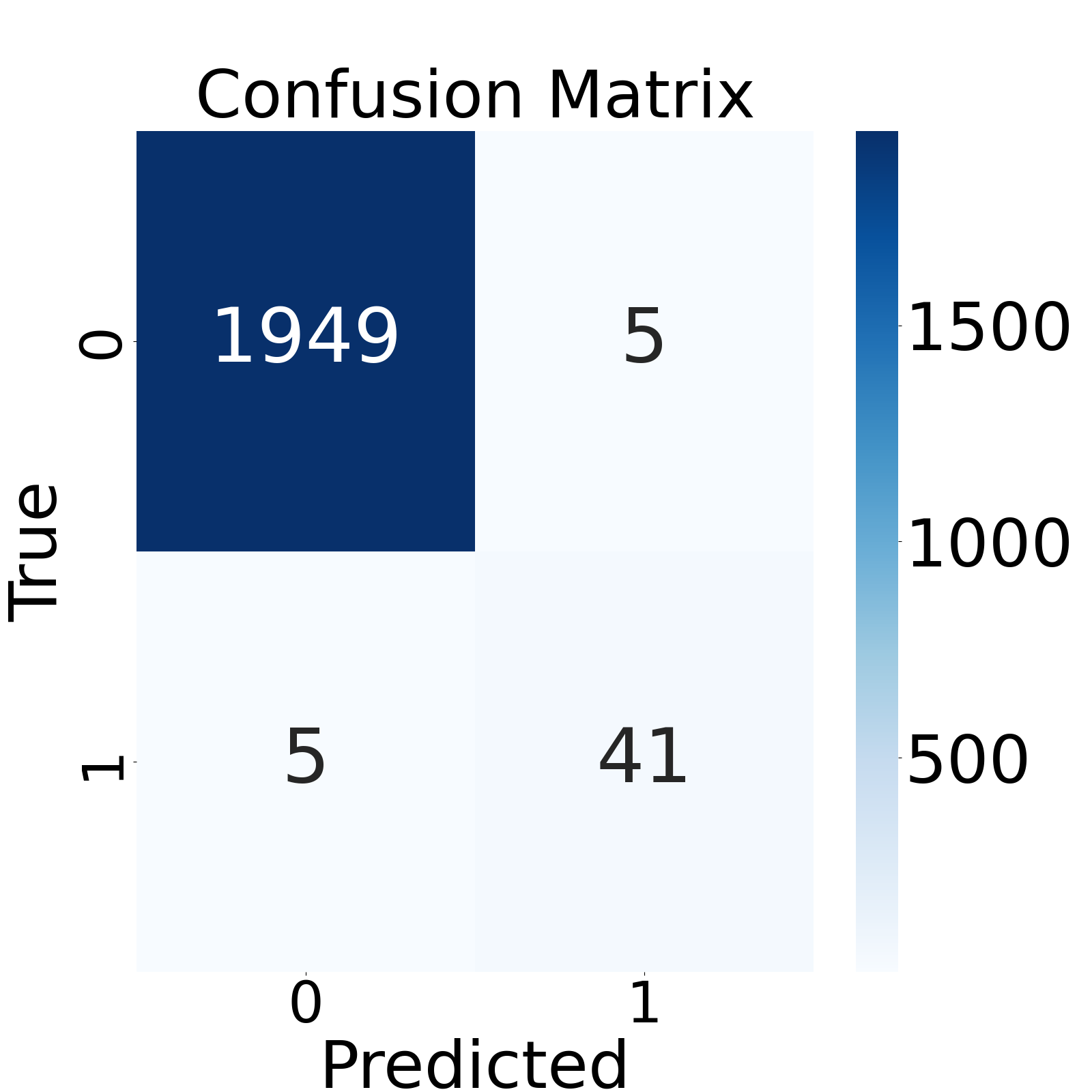}
        \caption{SVTN(1)}
    \end{subfigure}
    \hfill 
    \begin{subfigure}[b]{0.49\linewidth}
        \includegraphics[width=\linewidth,height=1.1in]{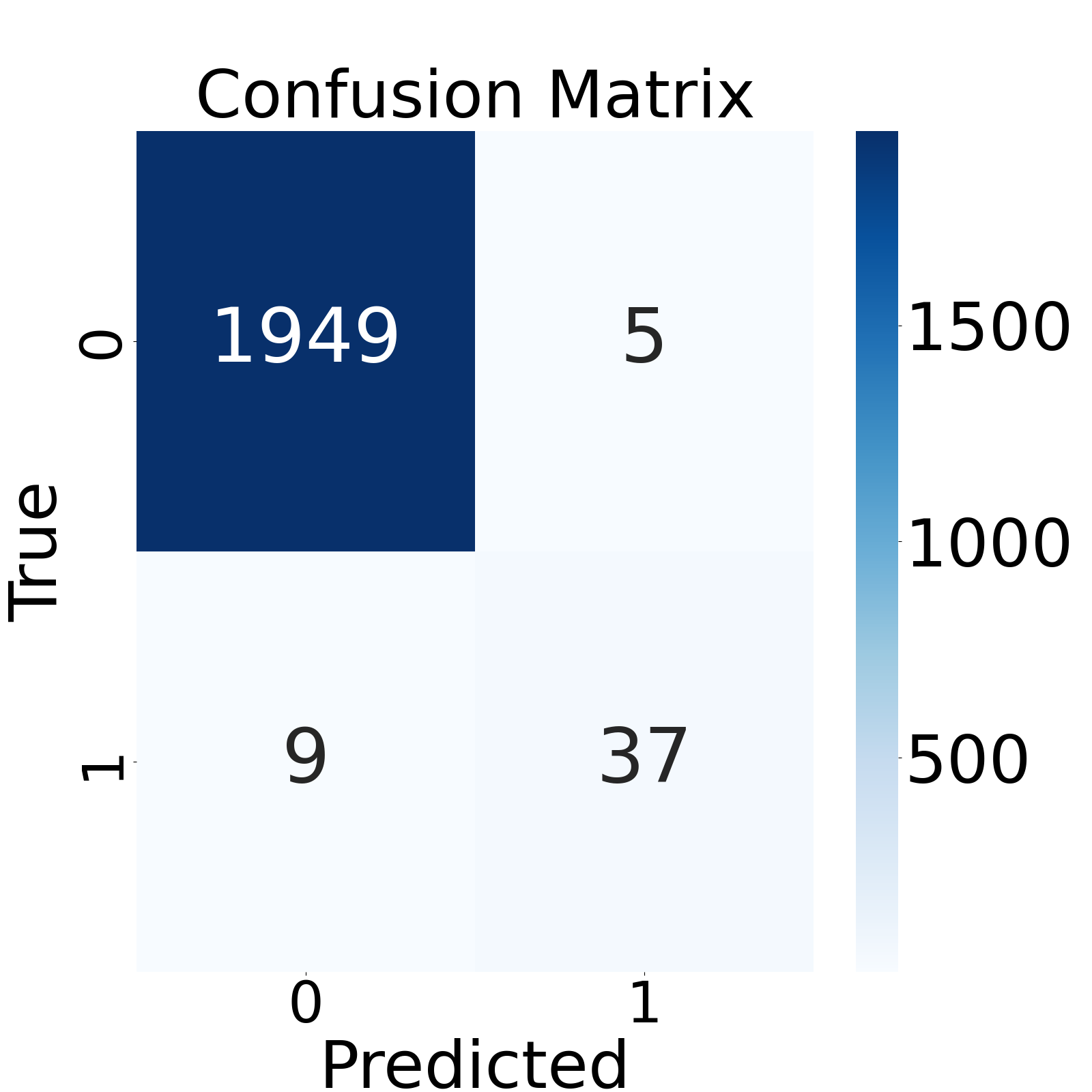}
        \caption{Transformer}
    \end{subfigure}
    
    \bigskip 
    \begin{subfigure}[b]{0.49\linewidth}
        \includegraphics[width=\linewidth,height=1.1in]{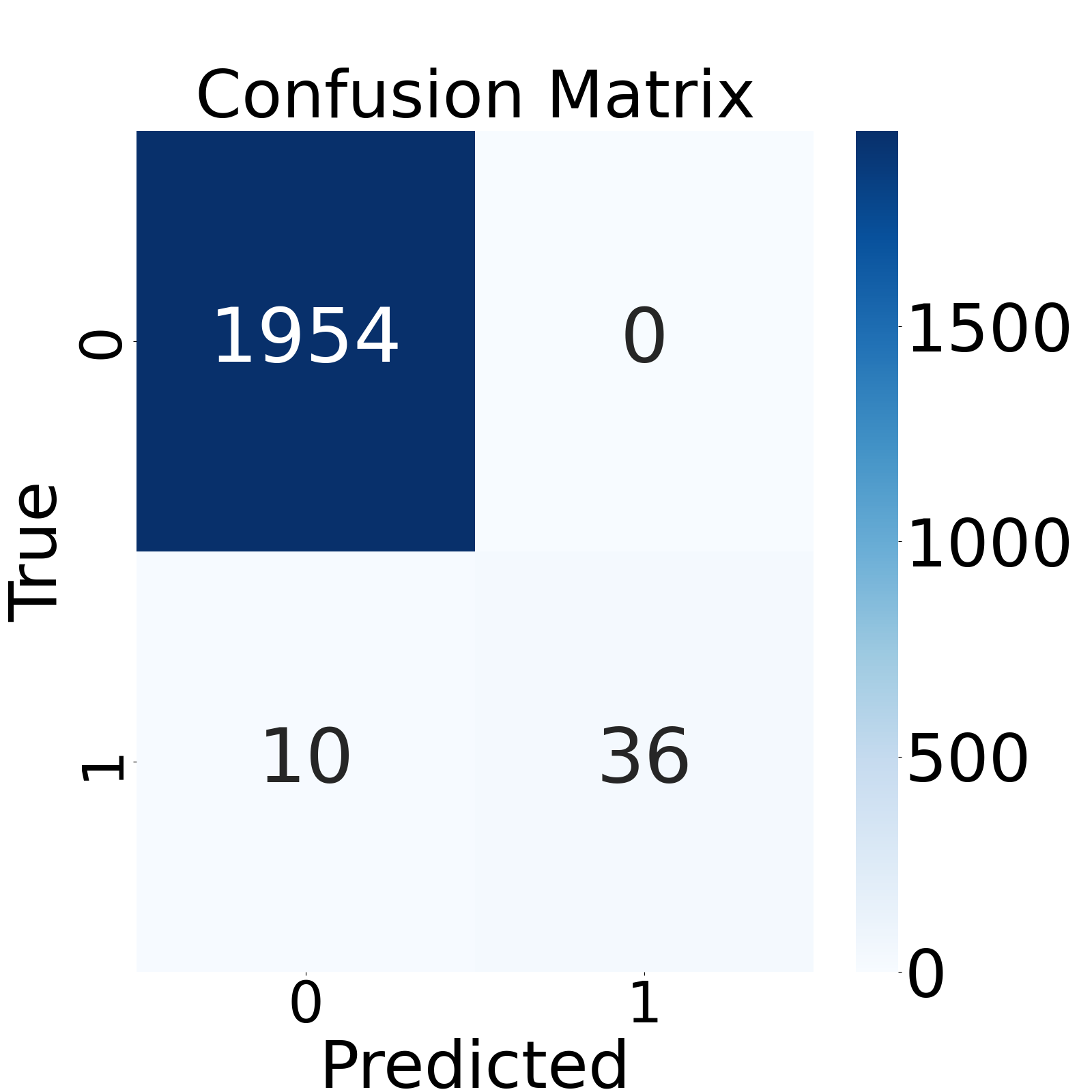}
        \caption{SV(5)}
    \end{subfigure}
    \hfill 
    \begin{subfigure}[b]{0.49\linewidth}
        \includegraphics[width=\linewidth,height=1.1in]{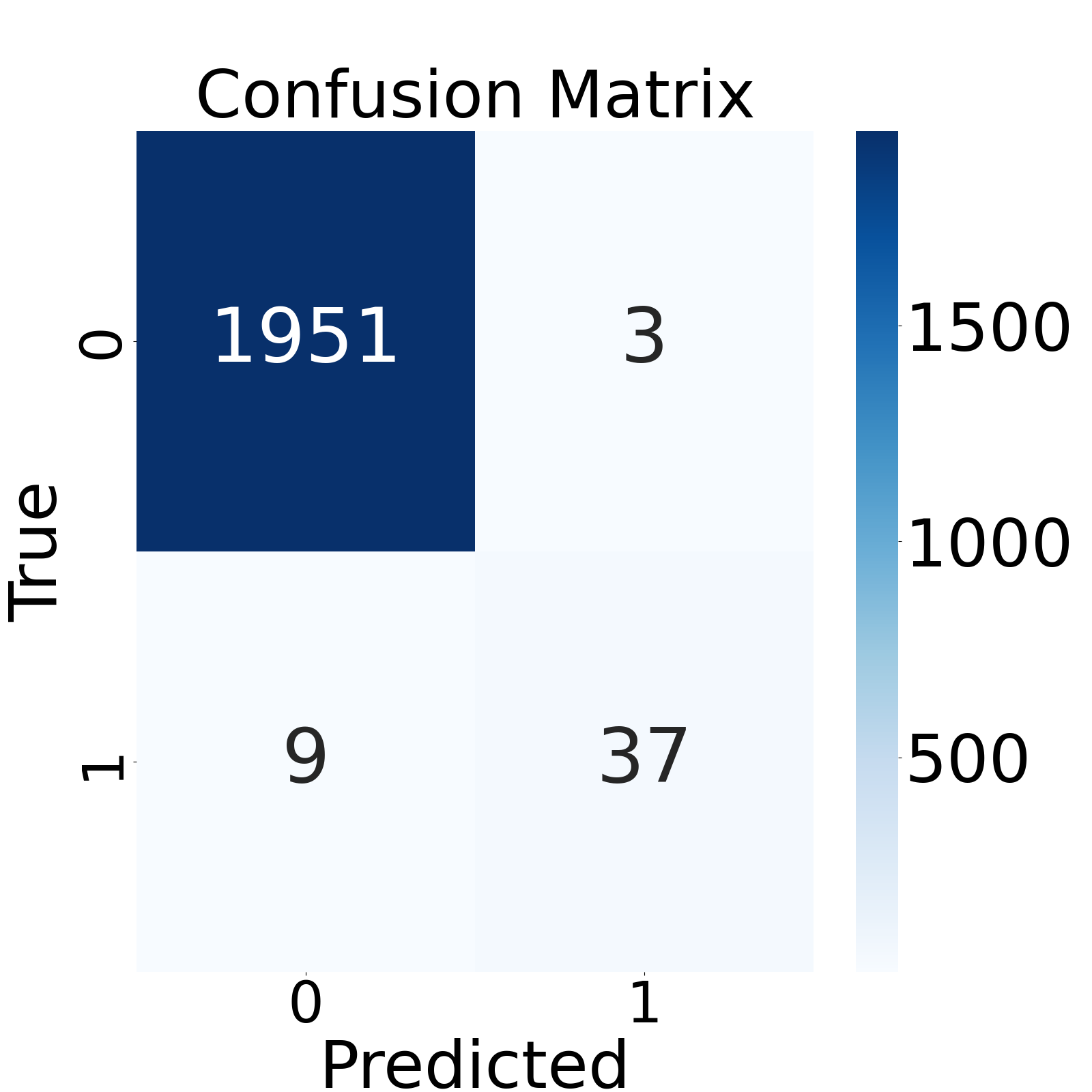}
        \caption{RF}
    \end{subfigure}
    \caption{Comparison of confusion matrics for four methods}
    \label{fig:cm}
\end{figure}

Fig.~\ref{fig:vcr} and Fig.~\ref{fig:cm} present the classification scatter plots and confusion matrices of the four methods, respectively. SVTN employs probabilistic modeling to establish the relationship between the transformer-generated feature vectors and the ground-truth labels, ensuring that the feature vectors converge more accurately toward the true labels. This innovative design endows the SVTN method with outstanding performance in emission identification tasks.

\begin{table}[tbp]
	\centering
	\caption{Comparison of performance metrics across SVTN parameters} 
	\scriptsize
	\setlength{\tabcolsep}{4pt}
	\begin{tabular}{@{} l *{4}{S[table-format=2.2(3)]} @{}}
		\toprule
		\multicolumn{1}{c}{\textbf{Metrics}} & \multicolumn{4}{c}{\textbf{Methods}} \\
		\cmidrule(lr){1-1} \cmidrule(lr){2-5}
		& {SVTN(1)} & {SVTN(3)} & {SVTN(5)} & {SVTN(7)} \\
		\midrule
		Recall$ (\uparrow)$ & $\bm{86.47 \pm 4.56}$ & 85.30(624) & 85.22(563) & 85.30(624) \\
		F1-score$ (\uparrow)$ & $\bm{89.16 \pm 3.42}$  & 88.35(417) & 88.00(347) & 88.35(417) \\
		\bottomrule
	\end{tabular}
	\label{tab:svtn}
\end{table}
\begin{figure}[!tbp]
	\centering
	\begin{subfigure}[b]{0.48\linewidth} 
		\includegraphics[width=\linewidth,keepaspectratio]{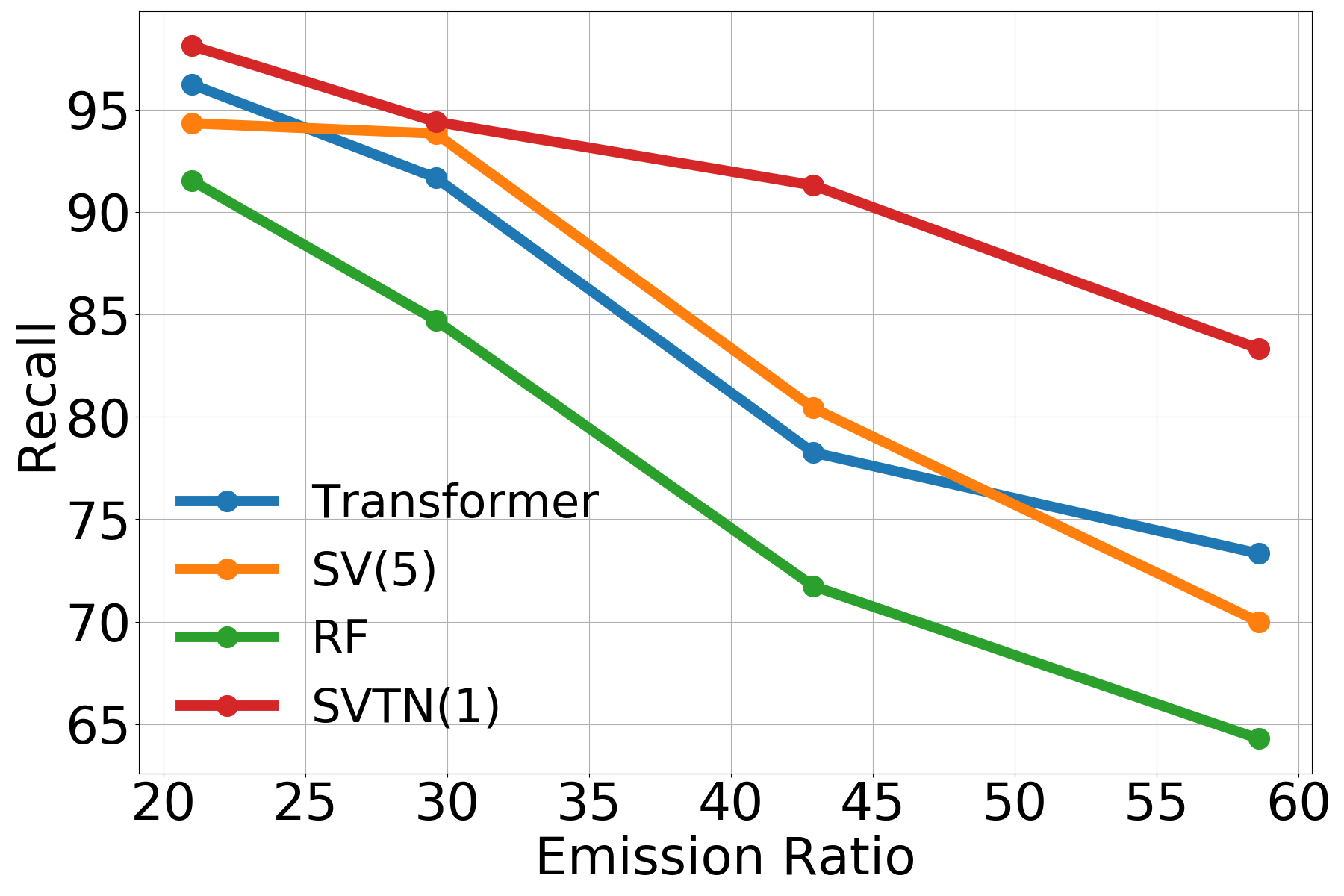}
		\caption{Recall}
	\end{subfigure}\hspace{-0.05cm} 
	\begin{subfigure}[b]{0.48\linewidth}
		\includegraphics[width=\linewidth,keepaspectratio]{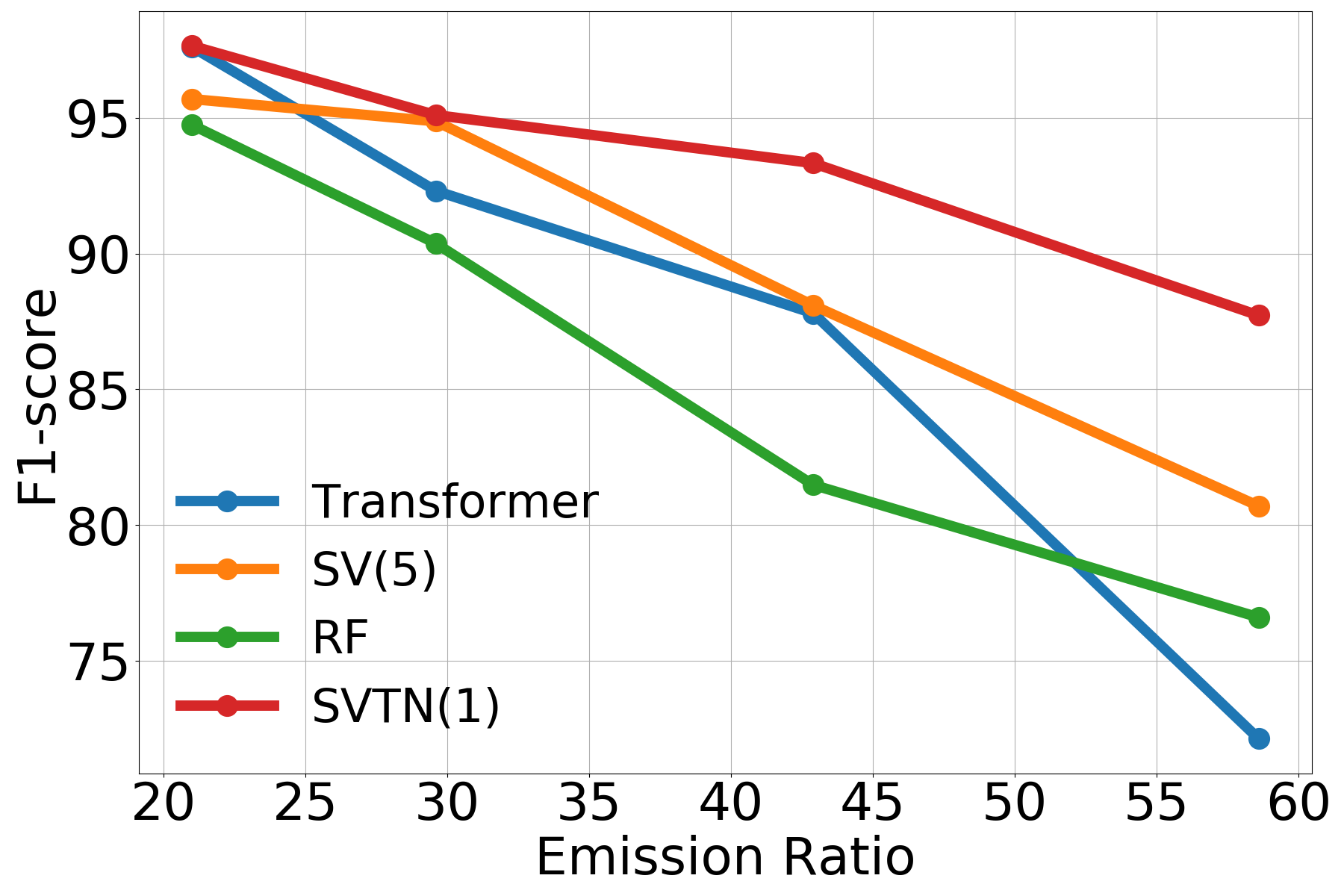}
		\caption{F1 Score}
	\end{subfigure}
	\caption{Comparison of recall and f1-Score for four methods under different Emission Ratios}
	\label{fig:xn}
\end{figure}
As shown in Tab.~\ref{tab:svtn}, the SVTN(1) model outperforms the high-order nonlinear SVTN method in terms of performance metrics, demonstrating the transformer's capability to extract features from raw data and map them into a low-dimensional space. To verify whether the SVTN model overcomes the dependency on data distribution in traditional data-driven algorithms, we adjusted the ratio of
normal-emission samples to high-emission samples (i.e., the Emission Ratio in Fig.~\ref{fig:xn}). The results indicate that as the Emission Ratio increases, the performance metrics of the SVTN model consistently remain superior to those of other models.

\section{CONCLUSIONS}\label{Section 5}

In response to the long-tailed distribution issue of actual emission data, this paper innovatively proposes the SVTN method. Unlike existing single-driver paradigm methods, SVTN leverages the transformer to deeply explore the evolution patterns of complex driving states of mobile sources on roads, providing prior knowledge for the SV model. The effectiveness of this method has been fully validated through evaluation on real vehicle OBD emission data from Hefei City.

\addtolength{\textheight}{-12cm}   









\bibliographystyle{IEEEtran}
\bibliography{root}

\begin{thebibliography}{10}
\providecommand{\url}[1]{#1}
\csname url@rmstyle\endcsname
\providecommand{\newblock}{\relax}
\providecommand{\bibinfo}[2]{#2}
\providecommand\BIBentrySTDinterwordspacing{\spaceskip=0pt\relax}
\providecommand\BIBentryALTinterwordstretchfactor{4}
\providecommand\BIBentryALTinterwordspacing{\spaceskip=\fontdimen2\font plus
\BIBentryALTinterwordstretchfactor\fontdimen3\font minus
  \fontdimen4\font\relax}
\providecommand\BIBforeignlanguage[2]{{%
\expandafter\ifx\csname l@#1\endcsname\relax
\typeout{** WARNING: IEEEtran.bst: No hyphenation pattern has been}%
\typeout{** loaded for the language `#1'. Using the pattern for}%
\typeout{** the default language instead.}%
\else
\language=\csname l@#1\endcsname
\fi
#2}}

\bibitem{KANG2021117877}
Y.~Kang, Z.~Li, W.~Lv, Z.~Xu, W.~X. Zheng, and J.~Chang, ``High-emitting
  vehicle identification by on-road emission remote sensing with scarce
  positive labels,'' \emph{Atmospheric Environment}, vol. 244, p. 117877, 2021.

\bibitem{10187171}
L.~Pei, Y.~Cao, Y.~Kang, Z.~Xu, and Z.~Zhao, ``Self-supervised spatiotemporal
  clustering of vehicle emissions with graph convolutional network,''
  \emph{IEEE Transactions on Neural Networks and Learning Systems}, vol.~35,
  no.~11, pp. 16\,301--16\,312, 2024.

\bibitem{Zhao2024ASL}
Z.-Y. Zhao, Y.~Cao, Z.~Xu, and Y.~Kang, ``A seq2seq learning method for
  microscopic emission estimation of on-road vehicles,'' \emph{Neural Comput.
  Appl.}, vol.~36, pp. 8565--8576, 2024.

\bibitem{xu2023high}
Z.~Xu, R.~Wang, Y.~Cao, and Y.~Kang, ``High-emitter identification for
  heavy-duty vehicles by temporal optimization {LSTM} and an adaptive dynamic
  threshold,'' \emph{Frontiers of Information Technology \& Electronic
  Engineering}, vol.~24, no.~11, pp. 1633--1646, 12 2023.

\bibitem{GUO2021479}
T.~GUO, N.~JIANG, B.~LI, X.~ZHU, Y.~WANG, and W.~DU, ``Uav navigation in high
  dynamic environments: A deep reinforcement learning approach,'' \emph{Chinese
  Journal of Aeronautics}, vol.~34, no.~2, pp. 479--489, 2021.

\bibitem{8946762}
Q.~Sun, K.~Zhang, and Y.~Shi, ``Resilient model predictive control of
  cyber–physical systems under dos attacks,'' \emph{IEEE Transactions on
  Industrial Informatics}, vol.~16, no.~7, pp. 4920--4927, 2020.

\bibitem{THALER202338125}
B.~Thaler, S.~Posch, A.~Wimmer, and G.~Pirker, ``Hybrid model predictive
  control of renewable microgrids and seasonal hydrogen storage,''
  \emph{International Journal of Hydrogen Energy}, vol.~48, no.~97, pp.
  38\,125--38\,142, 2023.

\bibitem{TU2023120289}
H.~Tu, S.~Moura, Y.~Wang, and H.~Fang, ``Integrating physics-based modeling
  with machine learning for lithium-ion batteries,'' \emph{Applied Energy},
  vol. 329, p. 120289, 2023.

\bibitem{BI20143220}
W.~Bi and Y.~Zhao, ``Iterative parameter estimate with batched binary-valued
  observations: Convergence with an exponential rate,'' \emph{IFAC Proceedings
  Volumes}, vol.~47, no.~3, pp. 3220--3225, 2014, 19th IFAC World Congress.

\bibitem{10383964}
J.~Guo, W.~Xue, T.~Wang, J.-F. Zhang, and Y.~Zhang, ``On iterative parameter
  identification of fir systems with batched possibly incorrect binary-valued
  observations,'' in \emph{2023 62nd IEEE Conference on Decision and Control
  (CDC)}, 2023, pp. 4936--4941.

\bibitem{GUO20133396}
J.~Guo and Y.~Zhao, ``Recursive projection algorithm on fir system
  identification with binary-valued observations,'' \emph{Automatica}, vol.~49,
  no.~11, pp. 3396--3401, 2013.

\bibitem{Li2022}
J.~Li, L.~Wu, W.~Lü, T.~Wang, Y.~Kang, D.~Feng, and H.~Zhou, ``Lithology
  classification based on set-valued identification method,'' \emph{Journal of
  Systems Science and Complexity}, vol.~35, no.~5, pp. 1637--1652, 10 2022.

\bibitem{6640899}
W.~Bi, Y.~Zhao, C.~Liu, and W.~Yue, ``Set-valued analysis for genome-wide
  association studies of complex diseases,'' in \emph{Proceedings of the 32nd
  Chinese Control Conference}, 2013, pp. 8262--8267.

\bibitem{10783042}
Y.~Xu, Y.~Zhang, and J.-F. Zhang, ``Parameter estimation and tracking control
  of mimo linear systems without prior knowledge of control signs and parameter
  bounds,'' \emph{IEEE Transactions on Automatic Control}, vol.~70, no.~6, pp.
  3695--3710, 2025.

\bibitem{cai2023statistical}
T.~T. Cai, Z.~Guo, and R.~Ma, ``Statistical inference for high-dimensional
  generalized linear models with binary outcomes,'' \emph{Journal of the
  american statistical association}, vol. 118, no. 542, pp. 1319--1332, 2023.

\bibitem{Ljung1998}
L.~Ljung, \emph{System Identification}.\hskip 1em plus 0.5em minus 0.4em\relax
  Boston, MA: Birkh{\"a}user Boston, 1998, pp. 163--173.

\bibitem{shao2008mathematical}
J.~Shao, \emph{Mathematical statistics}.\hskip 1em plus 0.5em minus 0.4em\relax
  Springer Science \& Business Media, 2008.

\bibitem{JMLR:v9:vandermaaten08a}
L.~van~der Maaten and G.~Hinton, ``Visualizing data using t-sne,''
  \emph{Journal of Machine Learning Research}, vol.~9, no.~86, pp. 2579--2605,
  2008.

\end{thebibliography}



\end{document}